\documentclass[letterpaper,10pt,conference]{IEEEtran}
\usepackage{amsmath,amssymb,booktabs,graphicx,microtype,cite}
\title{From Language to Motion: Task-Conditioned Focal-Stack Trajectory Integration for Microscopic Robots}
\author{%
\IEEEauthorblockN{Junjie Xie$^{1,\dagger}$, Chuxuan He$^{2,\dagger}$, Junkai Huang$^{1,\dagger}$, Heng Zhang$^{3,\dagger}$,\\
Angen Ye$^{1}$, Yujia Song$^{1}$, Yuqing Li$^{1}$, Pengsong Zhang$^{4}$, and Dapeng Zhang$^{1,*}$}%
\IEEEauthorblockA{$^{1}$\,Institute of Automation, Chinese Academy of Sciences\\
$^{2}$\,Zhejiang Gongshang University\\
$^{3}$\,Safe AI Lab, Carnegie Mellon University\\
$^{4}$\,Department of Mechanical and Industrial Engineering, University of Toronto,\\
Toronto, Ontario, M5S 3G8, Canada (\texttt{pengsong.zhang@mail.utoronto.ca})\\
$^{\dagger}$\,These authors contributed equally to this work.\\
$^{*}$\,Corresponding author: Dapeng Zhang.}%
}
\begin{document}
\maketitle
\noindent{\footnotesize\itshape This work has been submitted to the IEEE for possible publication. Copyright may be transferred without notice, after which this version may no longer be accessible.\par}
\smallskip
\thispagestyle{empty}
\pagestyle{empty}
\begin{abstract}
Microscopic visual servoing requires a robot to follow fine geometric features that a single image may reveal only in part. At high magnification, the field of view is limited, and shallow depth of field can leave different sections of the same contour sharp at different focal settings. These constraints also make training data costly to annotate, particularly when a change in part or task calls for a different edge, boundary, or offset path. We present a language-guided framework that addresses both the cost of configuring new tasks and the loss of usable geometry in any single focal plane. A language instruction specifies the geometry to recover, while frozen perception models and geometric operators locate it in the images. This separation enables zero-shot reconfiguration for the evaluated tasks without new task-specific training labels. To recover a path across focal planes, we extract local trajectories first and then select a consistent path from the segments supported by the image evidence. Each focal plane can thus contribute the sections it resolves well. The evaluated configuration took 15 minutes. Across nine part and illumination conditions, trajectory-level integration reduced 2-D contour RMSE by 56.4\% relative to image-first fusion with SAMFF. In robot dispensing experiments, target-region coverage increased from 83.5\% to 92.9\%.
\end{abstract}


\begin{figure*}[!t]
\centering
\includegraphics[width=\textwidth]{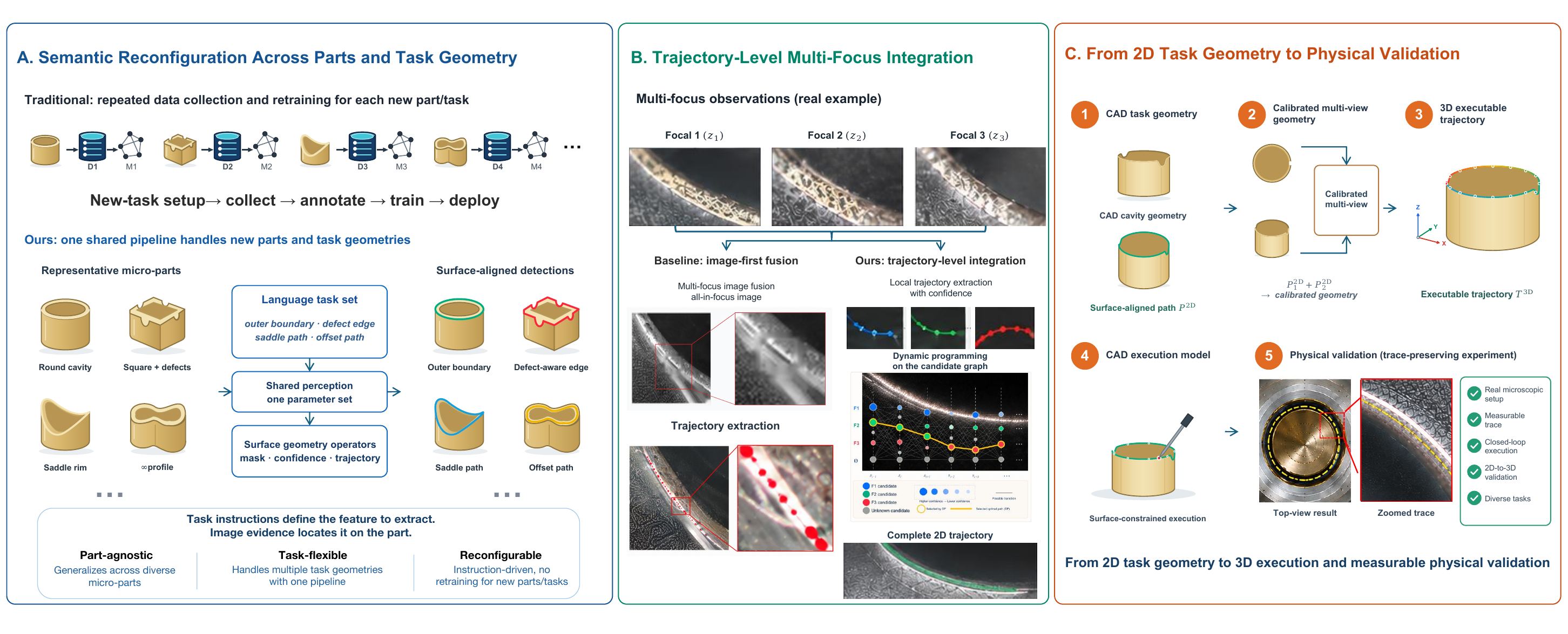}
\caption{System overview. (A) CAD-rendered examples illustrate task-level geometric reconfiguration through language prompts. (B) Trajectory candidates and local evidence are extracted at individual focal planes before integration in path space. (C) Calibrated multi-view geometry bridges the recovered 2-D paths to physical execution on the microscopic robot.}
\label{fig:fig1}
\end{figure*}

\section{Introduction}
\label{sec:introduction}

A microscopic robot must follow the geometry of its task with precision. In probing, contour following, or local material deposition, an error in locating an edge can change where the tool makes contact or where material is deposited. Yet the magnification needed to resolve that edge limits the field of view, and shallow depth of field can leave neighboring sections of a three-dimensional contour sharp at different focal settings. Recovering an ordered trajectory for execution therefore requires combining local observations of the geometry specified by the task. Fig.~\ref{fig:fig1} follows this problem from the language instruction through to the physical trace.

The same imaging constraints make it laborious to build a dataset for each task. An annotator tracing a contour may need to consult several focal images to identify its full extent. When a new part is introduced, or the task changes from following an outer boundary to following a local edge or an offset path, the required labels can change as well. Part-specific models such as U-Net and nnU-Net~\cite{ronneberger2015unet,isensee2021nnunet} can achieve high accuracy, but repeated data collection, annotation, and training add setup time in high-mix, low-volume work~\cite{sha2019review}. We use a constrained language specification to identify the object, region, and geometric operation. Frozen open-vocabulary perception~\cite{liu2024groundingdino,kirillov2023sam,ren2024groundedsam} and geometry operators then recover the requested feature from the images. The evaluated tasks can be reconfigured through the instruction and operator settings without collecting new training labels or retraining the visual backbone.

Across a focal stack, each image may offer a useful measurement of only part of the path~\cite{nayar1994shape,pertuz2013focus}. Multi-focus image fusion~\cite{bouzos2019crf,liu2020mfifsurvey} combines these images into an all-in-focus composite before geometry is extracted. At a focus transition, however, a composite can appear sharp while the boundary needed for servoing becomes displaced or ambiguous. We extract candidate paths from the individual focal planes and combine them directly in trajectory space. Local confidence determines how much weight each observation carries, and dynamic programming selects a geometrically consistent sequence along the path. This lets the system use well-resolved sections from different focal planes and leave an interval unresolved when the available evidence is too weak.

We recover the 2-D trajectory separately in each calibrated view and use standard multi-view geometry to reconstruct a 3-D path for the robot. The evaluation measures the two contributions separately. Changes in wording, part identity, and requested geometry test whether the same perception pipeline can be reconfigured with little setup. A matched comparison with SAMFF and component ablations test whether integrating focal evidence in trajectory space improves contour localization. Finally, a dispensing experiment leaves a measurable trace, allowing us to compare how closely the executed motion follows the target region.

\section{Related Work}
\label{sec:related}

\subsection{Language-Grounded Geometry Extraction}

U-Net and nnU-Net are supervised segmentation baselines~\cite{ronneberger2015unet,isensee2021nnunet}. Reusable segmentation has also been demonstrated by Cellpose~\cite{stringer2021cellpose} and microscopy-adapted foundation models~\cite{ma2024medsam,archit2025microSAM}. Language--image pretraining~\cite{radford2021clip} enables text-driven perception: Grounding DINO detects described objects~\cite{liu2024groundingdino}, SAM estimates masks~\cite{kirillov2023sam}, and Grounded SAM composes them~\cite{ren2024groundedsam}. Other text-conditioned interfaces include OWLv2 for box detection~\cite{minderer2023owlv2} and SAM~3 and SEEM for segmentation~\cite{carion2025sam3,zou2023seem}. Robot systems have also used natural language to structure downstream planning~\cite{ichter2023saycan,huang2023voxposer}. Here language specifies \emph{what geometry} to measure, not motor actions or metric locations; calibrated visual processing is still responsible for \emph{where} it lies. We do not propose a new visual foundation model.

\subsection{Focus Variation and Task-Space Fusion}

Shape-from-focus recovers depth cues from focal variation~\cite{nayar1994shape,pertuz2013focus}. Multi-focus image fusion (MFIF) instead synthesizes a composite image~\cite{liu2020mfifsurvey}. Focus transitions and image blending may displace edges used for downstream measurements~\cite{ma2020alphamatte,zhao2026edgepreserve}. Recent semantic or task-driven image fusion includes SAMFF~\cite{wei2026samff} and TDFusion~\cite{bai2025tdfusion}, but a fused image remains the output variable. We target an ordered geometric path directly: each focal plane contributes local hypotheses, and a path-level optimizer selects a consistent solution. The ability to leave an interval unspecified rather than force an unreliable prediction is related to selective prediction~\cite{geifman2019selectivenet}.

\subsection{Perception-to-Execution Bridge}

Multi-view reconstruction and vision-based robot control provide the established mathematical bridge from image measurements to robot motion~\cite{hartley2004multiview,chaumette2006visualservo}. Tracking quality is sensitive to upstream measurement uncertainty~\cite{kyrki2006measurement}, particularly in high-precision micromanipulation~\cite{tamadazte2010mems}. Physical traces can be measured against a registered desired path, as in deposition metrology~\cite{yamabe2024dispensing}. Our methodological focus is the upstream 2-D geometry passed to these established components. We report image-space trajectory errors and, separately, downstream target-relative physical-trace outcomes to evaluate both perception and execution.

\section{Method}
\label{sec:method}

Our method follows the same three-stage structure introduced above: language-guided 2-D task-geometry extraction, trajectory-level focal integration, and a standard 2-D-to-3-D execution bridge. The first two stages are the methodological focus; the third only connects the recovered geometry to the robot.

\subsection{Language-Guided 2-D Task Geometry}

\begin{figure*}[!t]
\centering
\includegraphics[width=\textwidth]{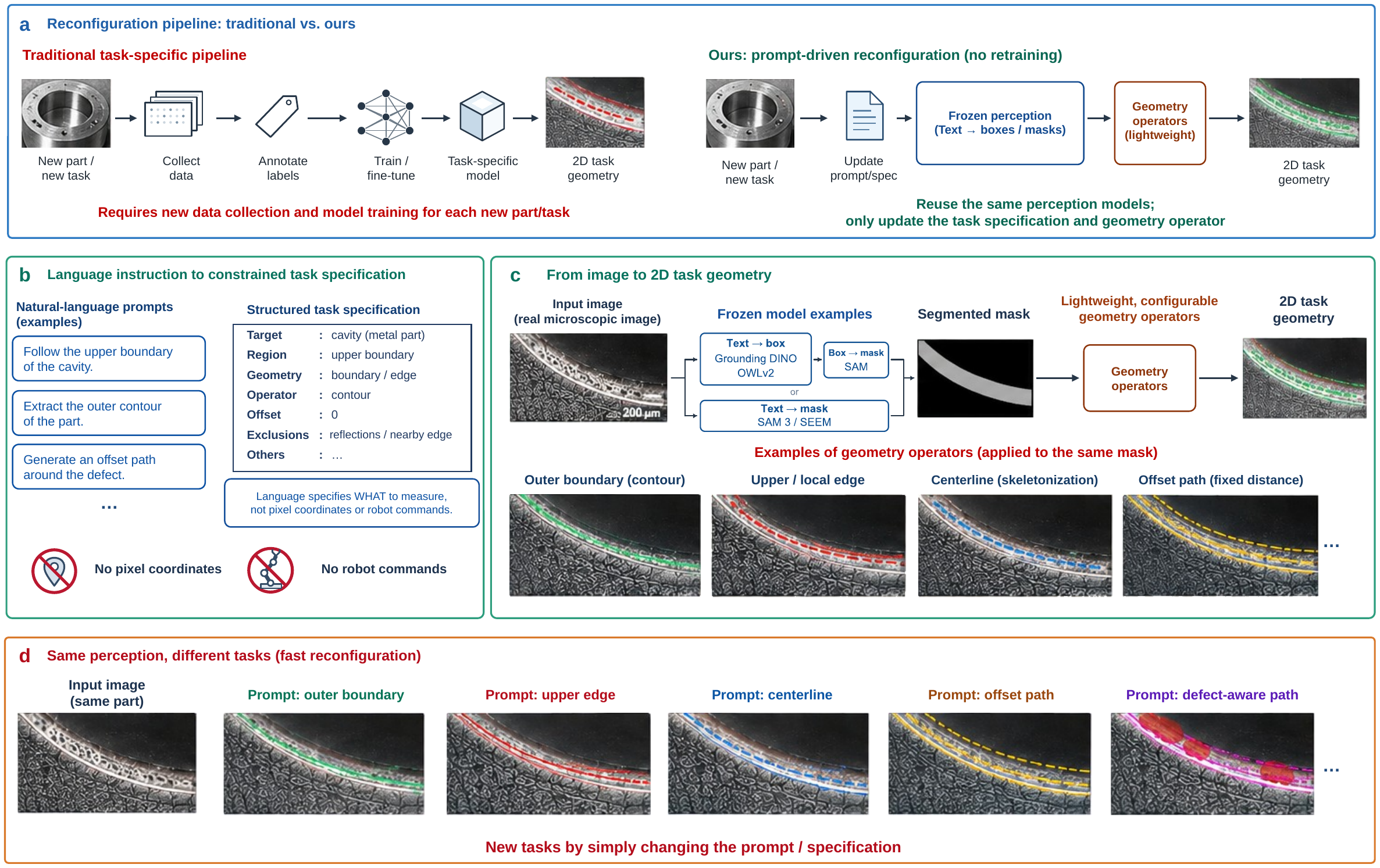}
\caption{Semantic-to-geometry reconfiguration. (a) Traditional part-specific retraining versus a prompt-driven workflow. (b) Natural-language instructions are converted into constrained task specifications. (c) Alternative frozen detector--segmenter or text-to-mask routes and lightweight geometry operators recover 2-D task geometry. (d) The same visual backbone serves multiple geometry requests without task-specific detector retraining.}
\label{fig:fig2}
\end{figure*}

Given a natural-language request $\ell$, a multimodal semantic model converts it into a constrained task specification
\begin{equation}
\mathcal S=(\ell^{+},\ell^{-},R,G,\theta_G),
\end{equation}
where $\ell^{+}$ and $\ell^{-}$ describe the desired and excluded targets, $R$ specifies the relevant region or spatial relation, $G$ denotes the requested geometric primitive, and $\theta_G$ contains lightweight task-specific settings. The semantic model specifies \emph{what} should be measured; it does not generate image coordinates, metric coordinates, or motor commands.

For camera view $v$ and focal plane $k$, a detector--segmenter realization produces
\begin{equation}
B_{v,k}=\operatorname{Det}(I_{v,k},\ell^{+},\ell^{-}), \qquad
M_{v,k}=\operatorname{Seg}(I_{v,k},B_{v,k}),
\end{equation}
using frozen general-purpose perception backbones. A frozen text-conditioned segmenter may alternatively produce $M_{v,k}$ directly. The named models in Fig.~\ref{fig:fig2} illustrate interface options. A task-specific post-processor then converts the mask into the requested 2-D geometry,
\begin{equation}
P_{v,k}=H_G(M_{v,k};\theta_G),
\end{equation}
where $H_G$ may contain geometric processing or a lightweight task-specific learning module. We do not require $H_G$ itself to be universal. The intended reconfiguration path is therefore $\ell\rightarrow\mathcal S\rightarrow P$, rather than rebuilding a full collect--annotate--train pipeline for every revised request. Figure~\ref{fig:fig2} illustrates the reusable perception and task-geometry interface.

\subsection{Confidence-Weighted Trajectory-Level Focal Integration}

\begin{figure*}[!t]
\centering
\includegraphics[width=\textwidth]{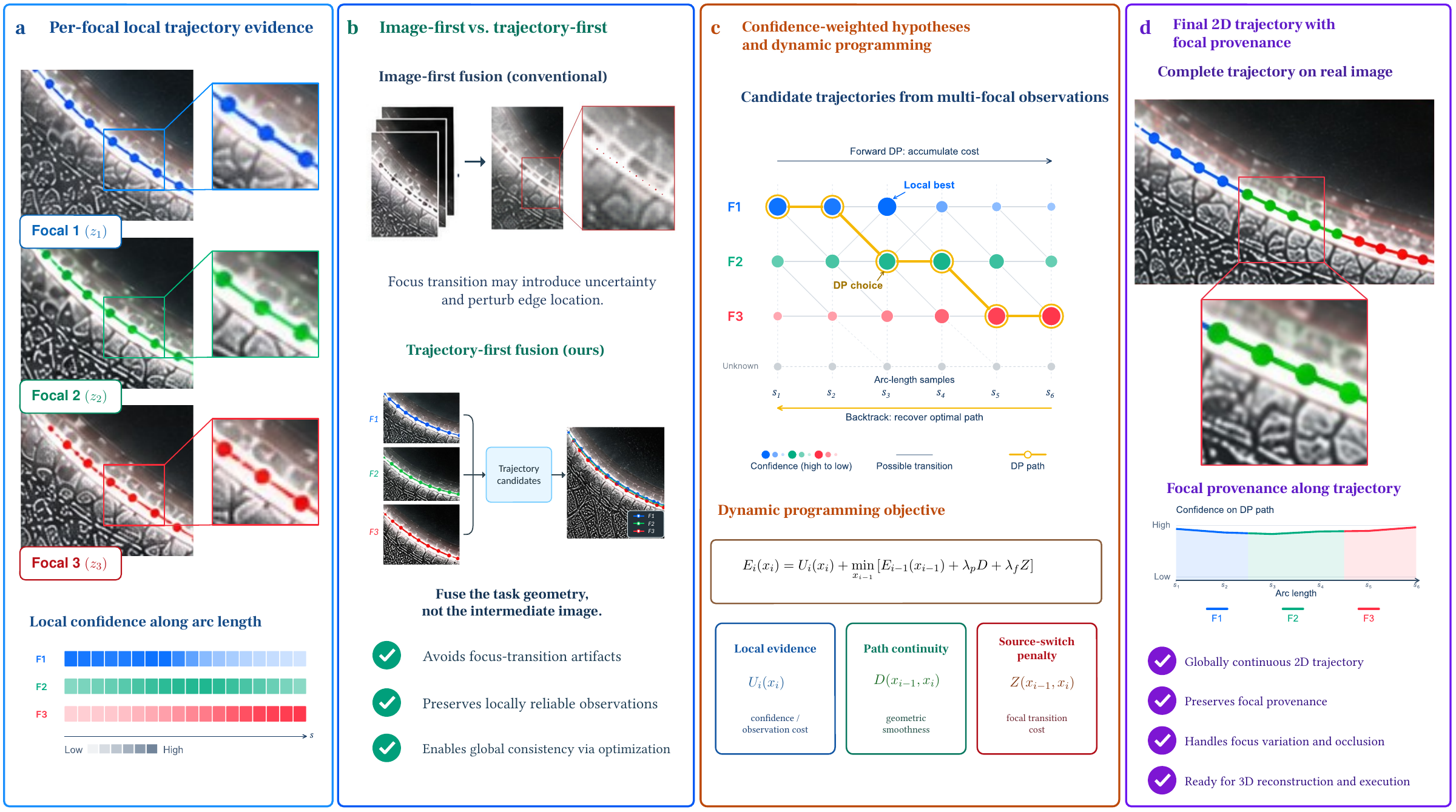}
\caption{Confidence-weighted trajectory-level multi-focal integration. (a) Local trajectories and reliability are estimated separately for each focal plane. (b) Image-first and trajectory-first processing paths. (c) Confidence-weighted hypotheses and dynamic-programming selection in ordered path space. (d) The final 2-D trajectory preserves focal-source information and confidence along the selected path. The confidence graphics are schematic.}
\label{fig:fig3}
\end{figure*}

Because shallow depth of field makes each focal plane reliable only over part of the desired trajectory, we avoid first collapsing the focal stack into an all-in-focus image. Instead,
\begin{equation}
\{I_k\}_{k=1}^{K}\rightarrow\{P_k,w_k\}_{k=1}^{K}\rightarrow P^{\ast},
\end{equation}
where $P_k$ is the trajectory extracted from focal plane $k$ and $w_k$ denotes nonnegative local confidence information; missing or rejected observations carry no weight.

Each candidate trajectory is resampled by normalized arc length $s\in[0,1]$. At location $s_i$, the candidate from focal plane $k$ is $\mathbf p_{k,i}=P_k(s_i)$ with confidence $w_{k,i}$. The confidence combines available task-postprocessing confidence, detector/segmenter confidence, local focus quality, and cross-focal geometric consistency. It is used only to compare local observations within the same task, not as a new universal reliability model.

Candidates that agree in position, local tangent, and neighborhood geometry are grouped into a local hypothesis $A_{i,h}$. Grouping is restricted to compatible observations so that conflicting boundaries are not simply averaged. Consistent observations can be confidence-weighted,
\begin{equation}
\bar{\mathbf p}_{i,h}=\frac{\sum_{k\in A_{i,h}}w_{k,i}\mathbf p_{k,i}}{\sum_{k\in A_{i,h}}w_{k,i}},
\end{equation}
whereas clearly separated edges remain competing hypotheses rather than being averaged.

Independent pointwise selection can cause focal-source flicker and geometric jumps. We therefore select the ordered trajectory globally with dynamic programming. Let $x_i$ denote the hypothesis selected at $s_i$, including a possible null
state $x_i=\varnothing$ when no observation is sufficiently supported. The
first-order, Viterbi-style dynamic-programming recurrence~\cite{viterbi1967dp} is
\begin{equation}
\begin{aligned}
E_i(x_i)={}&U_i(x_i)+\min_{x_{i-1}}\bigl[ E_{i-1}(x_{i-1})\\
&+\lambda_p D(x_{i-1},x_i)\\
&+\lambda_f Z(x_{i-1},x_i)\bigr],
\end{aligned}
\end{equation}
where $U_i$ is the local evidence cost (including a cost for the null state), $D$ penalizes spatial discontinuity, and $Z$ discourages unsupported focal-source switching. Backtracking yields the accepted portions of the fused per-view 2-D trajectory $P_v^{\ast}=\{\mathbf u_{v,i}\}$. Figure~\ref{fig:fig3} shows the local evidence, competing hypotheses, and global selection process.

\subsection{From 2-D Task Geometry to 3-D Execution}

The above inference is applied independently in every calibrated camera view. Corresponding 2-D trajectories are then lifted by standard calibrated multi-view geometry into an executable 3-D task trajectory $\mathcal T^{3D}=\{\mathbf q_i\}$. A generic formulation is
\begin{equation}
\hat{\mathbf q}_i=\arg\min_{\mathbf q}\sum_v c_{v,i}\|\pi_v(\mathbf q)-\mathbf u_{v,i}\|_2^2,
\end{equation}
where $\pi_v$ is the calibrated projection model. The geometric reconstruction follows established multi-view formulations~\cite{hartley2004multiview}. Our validation platform uses two approximately orthogonal microscopic views as one practical configuration; orthogonality is not required by the method, and focal-plane indices are not treated as metric depth.

The reconstructed 3-D trajectory is supplied to the robot control interface. We evaluate the integration algorithm against manually annotated 2-D paths and characterize the physical outcome using target-relative trace metrics after execution. The two metric families are reported separately to avoid conflating image-space localization and robot-level accuracy.

\section{Experiments}
\label{sec:experiments}

\begin{figure*}[!t]
\centering
\includegraphics[width=\textwidth]{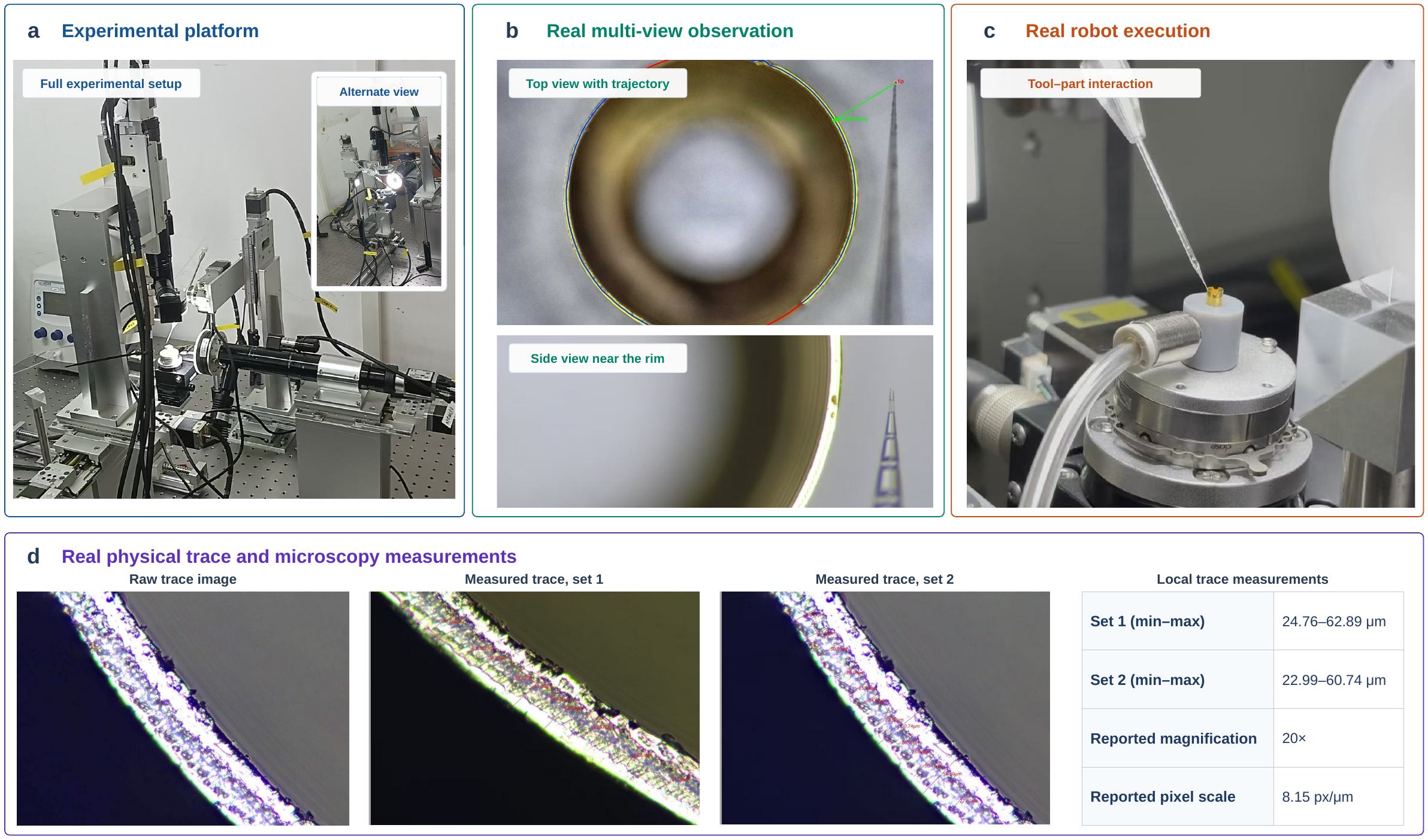}
\caption{Real-system physical validation. (a) Microscopic manipulation platform and auxiliary setup. (b) Complementary microscope views. (c) Robot--part interaction during execution. (d) Deposited physical traces with local trace measurements and their min--max ranges. Dispensing is used as a trace-preserving measurement task; target-relative performance is summarized separately in Table~\ref{tab:physical_demo}.}
\label{fig:fig4}
\end{figure*}

We evaluate three linked questions: (A) whether language, part, and requested-geometry changes preserve trajectory accuracy at low adaptation cost; (B) whether integration of focal evidence in trajectory space improves localization, and which components contribute; and (C) whether the resulting 3-D execution pipeline gives measurable target-aligned physical traces. Two-dimensional errors are compared with manually annotated reference trajectories, whereas physical metrics characterize the deposited trace after execution.

\subsection{Language Generalization and Reconfiguration}
\label{sec:exp_language}

Figure~\ref{fig:fig2} shows how the language interface maps a request to a constrained geometric operator while reusing the same foundation perception components. We vary instruction wording, part identity, and the requested geometric primitive, with the remaining geometry-recovery pipeline held fixed. Table~\ref{tab:language_demo} reports the resulting image-space RMSE and 95th-percentile point error relative to manual annotations. The $\pm$ values are standard deviations across distinct inputs in each condition. The canonical, formal, semantic, and colloquial phrasings probe language sensitivity; parts D--F assess previously unseen part identities; and the local-edge, upper-envelope, and offset requests exercise distinct geometric outputs.
\begin{table}[t]
\caption{Language and task reconfiguration.}
\label{tab:language_demo}
\centering
\scriptsize
\setlength{\tabcolsep}{3.4pt}
\begin{tabular}{@{}lcc@{}}
\toprule
Condition & RMSE (px) $\downarrow$ & P95 (px) $\downarrow$\\
\midrule
Canonical wording & $6.30\pm0.03$ & $8.08\pm0.03$\\
Formal paraphrase & $6.34\pm0.03$ & $8.12\pm0.04$\\
Semantic paraphrase & $6.43\pm0.03$ & $8.25\pm0.03$\\
Colloquial wording & $6.57\pm0.03$ & $8.37\pm0.06$\\
\midrule
Unseen part D & $6.39\pm0.03$ & $8.19\pm0.03$\\
Unseen part E & $6.50\pm0.03$ & $8.26\pm0.04$\\
Unseen part F & $6.59\pm0.03$ & $8.37\pm0.05$\\
\midrule
Local edge & $6.37\pm0.03$ & $8.15\pm0.03$\\
Upper envelope & $6.47\pm0.03$ & $8.25\pm0.05$\\
Offset trajectory & $6.56\pm0.03$ & $8.36\pm0.06$\\
\bottomrule
\end{tabular}
\end{table}

Across the four phrasings, RMSE changes from 6.30 to 6.57 pixels. The three unseen parts yield 6.39--6.59 pixels, while the geometric requests yield 6.37--6.56 pixels. The comparatively limited variation indicates that the same perception interface can handle these controlled changes without rebuilding the visual backbone. These conditions bound the evidence for reconfiguration; they do not imply unrestricted open-set performance.

We additionally compare against U-Net models adapted with 20, 50, or 100 part-specific training annotations. The annotation count in Table~\ref{tab:unet_demo} refers to \emph{new adaptation labels}, not the manual references used to score the methods. Adaptation time is reported in minutes for the respective configurations, while trajectory accuracy is evaluated using the same pixel-space criteria. The comparison uses the test images and manual reference trajectories from Table~\ref{tab:focal_real}. U-Net times cover training. Our time covers configuration. Training and test sets were disjoint.
\begin{table}[t]
\caption{Task-specific supervision versus reconfiguration cost.}
\label{tab:unet_demo}
\centering
\scriptsize
\setlength{\tabcolsep}{2.7pt}
\begin{tabular}{@{}lcccc@{}}
\toprule
Method & Labels & RMSE & P95 & Adapt. time\\
\midrule
U-Net (100) & 100 & $5.47$ & $7.15$ & $165$ min\\
U-Net (50) & 50 & $5.78$ & $7.52$ & $119$ min\\
U-Net (20) & 20 & $6.46$ & $8.26$ & $72$ min\\
Ours & 0 & $6.28$ & $8.13$ & $15$ min\\
\bottomrule
\end{tabular}
\end{table}

The 100-label U-Net achieves the lowest RMSE (5.47 pixels), but takes 165 min to adapt; the proposed configuration attains 6.28 pixels with no new part-specific training labels and 15 min of configuration time. This corresponds to 150 min less adaptation effort in the measured comparison, at a cost of 0.81 pixels RMSE relative to the 100-label model. Against the 20-label U-Net, our approach is both quicker (15 versus 72 min) and more accurate (6.28 versus 6.46 pixels). The result is an \emph{accuracy--adaptation-cost} trade-off, not a claim that frozen perception always outperforms supervised training.

\subsection{Trajectory-Level Focal-Stack Integration}
\label{sec:exp_focal}

The focal-stack study uses three microscopic parts under three illumination settings, yielding nine matched part--illumination conditions. The image-first baseline, SAMFF~\cite{wei2026samff}, produces an all-in-focus image before trajectory extraction; our method extracts trajectory candidates at each focal plane and integrates them with confidence and dynamic programming. Each method is evaluated against the same manually annotated reference path in each condition. For sampled points $\hat{\mathbf p}_j$ and reference points $\mathbf g_j$, the pointwise distance is $e_j=\|\hat{\mathbf p}_j-\mathbf g_j\|_2$. We compute $\mathrm{RMSE}=\sqrt{N^{-1}\sum_j e_j^2}$, the 95th-percentile error (P95), mean error, and maximum error. Table~\ref{tab:focal_real} reports the mean and standard deviation across the nine conditions, rather than an independent-trial confidence interval. Tables~\ref{tab:unet_demo} and~\ref{tab:ablation_demo} reuse our RMSE and P95 means from these nine conditions.
\begin{table}[t]
\caption{Measured trajectory error across nine part--illumination conditions.}
\label{tab:focal_real}
\centering
\scriptsize
\setlength{\tabcolsep}{3.2pt}
\begin{tabular}{@{}lcccc@{}}
\toprule
Method & RMSE & P95 & Mean & Max\\
\midrule
SAMFF & $14.41\pm0.04$ & $20.07\pm0.09$ & $13.50\pm0.07$ & $20.09\pm0.06$\\
Ours & $\mathbf{6.28\pm0.04}$ & $\mathbf{8.13\pm0.05}$ & $\mathbf{5.93\pm0.05}$ & $\mathbf{9.25\pm0.05}$\\
Reduction & 56.4\% & 59.5\% & 56.1\% & 54.0\%\\
\bottomrule
\end{tabular}
\end{table}

The trajectory-first method reduces average RMSE by 56.4\% (14.41 to 6.28 pixels) and P95 by 59.5\% (20.07 to 8.13 pixels) against the image-first baseline. Its RMSE spans 6.23--6.36 pixels across conditions, compared with 14.34--14.47 pixels for SAMFF; the corresponding P95 ranges are 8.06--8.22 and 19.94--20.23 pixels. These are \emph{2-D image-space} metrics, distinct from physical-trace error.

To separate the effects of local reliability and global continuity, we compare pointwise maximum selection, confidence weighting without global DP, DP selection without confidence weighting, and the complete method under identical external conditions. Table~\ref{tab:ablation_demo} reports geometric error and focal-source switching, which measures how frequently the selected focal source changes along the recovered path.
\begin{table}[t]
\caption{Confidence/DP ablation.}
\label{tab:ablation_demo}
\centering
\scriptsize
\setlength{\tabcolsep}{3.4pt}
\begin{tabular}{@{}lccc@{}}
\toprule
Variant & RMSE & P95 & Source switches\\
\midrule
Pointwise max & $7.00$ & $8.81$ & $60.6$\\
Confidence only & $6.75$ & $8.53$ & $48.8$\\
DP only & $6.56$ & $8.33$ & $11.2$\\
Confidence + DP & $\mathbf{6.28}$ & $\mathbf{8.13}$ & $9.0$\\
\bottomrule
\end{tabular}
\end{table}

Confidence-only integration improves RMSE from 7.00 to 6.75 pixels relative to pointwise selection, while DP-only selection yields 6.56 pixels and cuts source switching from 60.6 to 11.2. Combining confidence with DP provides the lowest RMSE (6.28 pixels), P95 (8.13 pixels), and switch count (9.0). The ablation supports both local reliability weighting and globally consistent hypothesis selection rather than assigning the entire improvement to either component alone.

\subsection{From 2-D Geometry to Physical Validation}
\label{sec:exp_physical}

For physical validation, the per-view 2-D paths are converted to a 3-D task trajectory through calibrated multi-view geometry and executed on the microscopic robot. The dispensing experiment is selected for the persistence and measurability of its deposited trace, not because the perception method requires a dispensing tool. The platform, microscope views, execution, and physical traces are shown in Fig.~\ref{fig:fig4}.

Table~\ref{tab:physical_demo} presents physical-trace metrics under the image-first and trajectory-first upstream estimates. Relative trace error is normalized to the image-first result (100\%). Coverage and out-of-region deposition are expressed as percentages of the respective target-region and trace evaluation quantities; they must not be conflated with pixel-space RMSE. Each method was evaluated in 100 independent trials. The values marked $\pm$ are standard deviations.
\begin{table}[t]
\caption{Representative physical validation.}
\label{tab:physical_demo}
\centering
\scriptsize
\setlength{\tabcolsep}{3.0pt}
\begin{tabular}{@{}lccc@{}}
\toprule
Upstream trajectory & \shortstack{Trace error\\(\% of baseline)} & Coverage & Out-of-region\\
\midrule
Image-first MFIF & $100\%$ & $83.5\pm1.1\%$ & $12.5\pm0.5\%$\\
Ours & $51.9\%$ & $92.9\pm1.1\%$ & $5.4\pm0.5\%$\\
\bottomrule
\end{tabular}
\end{table}

The relative physical trace error decreases from 100\% to 51.9\%, a 48.1\% reduction after downstream reconstruction and execution. Target-region coverage increases by 9.4 percentage points (83.5\% to 92.9\%), while out-of-region deposition falls by 7.1 points (12.5\% to 5.4\%). As expected for an end-to-end system, the normalized physical improvement need not equal the upstream 2-D RMSE ratio ($6.28/14.41\approx0.436$); camera calibration, reconstruction, feedback, mechanism behavior, and material deposition all affect the executed trace.

Figure~\ref{fig:fig4}(d) shows local trace measurements with a reported 20$\times$ magnification, a source-image pixel scale of 8.15 px/$\mu$m, and min--max ranges of 24.76--62.89 and 22.99--60.74 $\mu$m. These local image annotations characterize the deposited features; the target-relative metrics in Table~\ref{tab:physical_demo} are separate evaluation quantities.

\subsection{Scope of the Evaluation}
\label{sec:exp_limitations}

The results cover the reported wording variants, three held-out part identities, three geometry requests, three U-Net adaptation budgets, nine part--illumination conditions, and one representative physical task. The study establishes the performance of the evaluated settings; larger part families, further MFIF algorithms, and different manipulation hardware would be required to generalize the numerical effect sizes beyond this scope.

\section{Conclusion}
\label{sec:conclusion}

We presented a language-conditioned microscopic task-geometry framework that decouples semantic requests from metric measurement, reuses frozen open-vocabulary perception, and integrates focal-plane trajectories directly in ordered path space through confidence weighting and dynamic programming. The comparison with part-specific U-Net models demonstrates a practical accuracy--adaptation-cost trade-off: the zero-new-label configuration achieves 6.28-pixel RMSE with 15 min of adaptation, whereas the most accurate 100-label U-Net obtains 5.47 pixels with 165 min of adaptation.

For nine matched part--illumination conditions, trajectory-space integration reduces RMSE and P95 by 56.4\% and 59.5\% relative to image-first MFIF. Component ablations further show that confidence and DP jointly reduce geometric error and focal-source switching. The physical trace experiment reports a 48.1\% reduction in normalized target-relative error, together with higher region coverage and reduced out-of-region deposition. These observations link task-level visual improvements to a real robotic outcome, while calibrated multi-view reconstruction remains a standard execution bridge rather than a new algorithmic contribution. Future work will test broader part families, focal-fusion baselines, and manipulation tasks beyond the trace-preserving validation used here.

\section*{Acknowledgment}
OpenAI Codex helped draft the Abstract and Introduction and prepare diagrams in Figs.~1--4.
\begingroup
\renewcommand{\baselinestretch}{0.95}

\endgroup
\end{document}